# Probabilistic Conceptual Network: A Belief Representation Scheme for Utility-Based Categorization

**Kim Leng Poh***
Laboratory for Intelligent Systems
Department of Engineering-Economic Systems
Stanford University, CA 94305-4025
poh@lis.stanford.edu

**Michael R. Fehling**
Laboratory for Intelligent Systems
Department of Engineering-Economic Systems
Stanford University, CA 94305-4025
fehling@lis.stanford.edu

## Abstract

*Probabilistic conceptual network* is a knowledge representation scheme designed for reasoning about concepts and categorical abstractions in utility-based categorization. The scheme combines the formalisms of abstraction and inheritance hierarchies from artificial intelligence, and probabilistic networks from decision analysis. It provides a common framework for representing conceptual knowledge, hierarchical knowledge, and uncertainty. It facilitates dynamic construction of categorization decision models at varying levels of abstraction. The scheme is applied to an automated machining problem for reasoning about the state of the machine at varying levels of abstraction in support of actions for maintaining competitiveness of the plant.

## 1 Introduction

A probabilistic conceptual network (pc-net) is a knowledge representation scheme designed to support utility-based categorization (Poh, 1993). In contrast to the traditional approaches which are logic and similarity-based (Smith & Medin, 1981), utility-based categorization considers the usefulness of the information conveyed by the concepts, the actional consequences, the desirability of the consequences of actions, the computational or cognitive resource requirement and availability, and the uncertainty about the environment.

We have developed a decision-theoretic approach for utility-based categorical reasoning as shown in Figure 1, in contrast to previous work on abstraction in probabilistic reasoning (Horvitz, Heckerman, Ng, & Nathwani, 1989; Horvtiz & Klein, 1992) which were more narrowly focused. In our view, a resource-constrained agent operating in an uncertain world is given a set of limited observations. It must conceptualizes the situation and decide on the most appropriate course of action. It does so by solving a categorization decision-model. However, different models at different levels of categorical abstraction can be used. Each of these models has different expected value of the recommended action and different computational resource requirement. The agent must therefore decide on the best level of abstraction to construct the model so as to achieve the best trade off between the expected value of the recommended action and cost of computation.

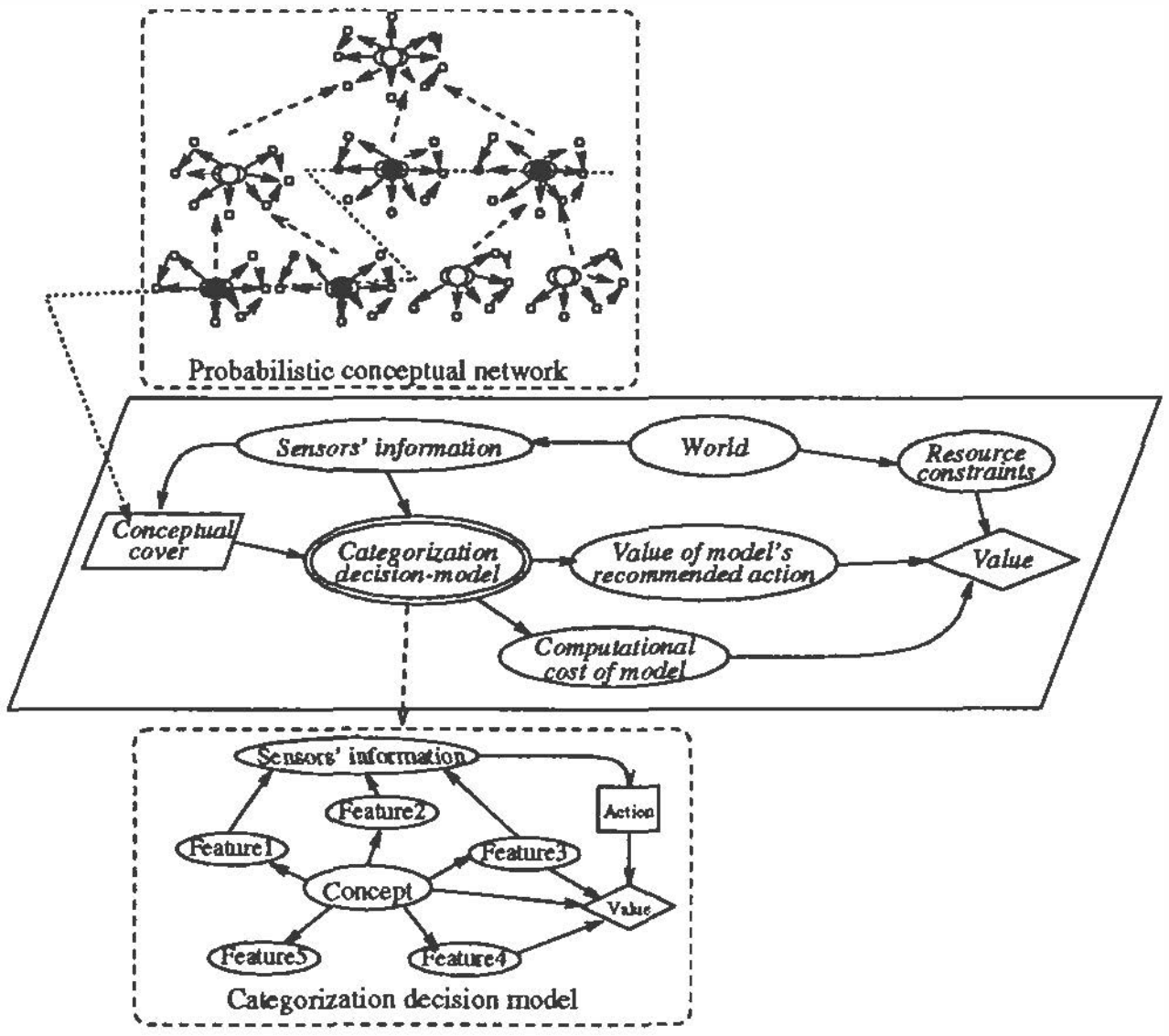

Figure 1: Using a pc-net in utility-based categorization

A probabilistic representation of conceptual categories called a pc-net is used to represent the agent's knowledge about the world. A level of *conceptual abstraction* for a building a model is obtained by selecting a *conceptual cover* from the pc-net. As illustrated in Figure 1, a conceptual cover is obtained by selecting a set of mutually exclusive and exhaustive concepts from different levels in the pc-net[1]

*Also at Dept. of Industrial & Systems Engineering, National University of Singapore, Kent Ridge, Singapore 0511

[1]The notion of conceptual coverage in abstraction hier-

We have developed an incremental algorithm whereby the reasoner iteratively specializes or generalizes the conceptual cover. A concept is *specialized* by breaking it up into a set of more specific subconcepts. A group of concepts may be *generalized* by replacing them with a single super-concept. At each iteration, changes are made in order to achieve the highest expected improvement in overall utility (Poh, 1993). The procedure applies the principles of decision-theoretic control (Horvitz, 1987, 1990; Fehling & Breese, 1987; Russell & Wefald, 1991) to iteratively decide between allocation of additional resources to refine the current set of concepts, or to act immediately based on the current action recommended by the model. This model refinement approach is a special application of a more general approach for refining general decision models (Poh & Horvitz, 1993).

In this paper, we describe *probabilistic conceptual networks* and show how they may be used to represent both categorical and uncertain knowledge and to facilitate the dynamic construction of categorization decision models at varying levels of abstraction. We present an example from automated machining. We also compare our scheme with similarity networks (Heckerman, 1991) and other approaches to knowledge-based decision model construction.

## 2 Integrating Uncertainty and Categorical Knowledge

To perform utility-based categorization, an intelligent actor must express different dimensions or perspectives of knowledge. First, she must be able to express categorical knowledge with some degree of modularity. Categorical knowledge expresses facts about individual concepts in a given domain, i.e., it describes the features and properties that characterize the concepts. Second, the actor must represent categorical relations, e.g, how one concept subsumes others. In particular, the actor, when problem-solving, must decide which concepts to use and at which levels of abstraction in order to obtain a useful solution.

In artificial intelligence, abstraction hierarchies and semantics nets (Lehmann, 1992) are graph-based formalisms that have been advocated for computer representation of concepts and categorical knowledge. They organize conceptual knowledge in levels of abstraction and make use of "inheritance" mechanisms whereby concepts may share features and properties with higher-level ones. Since feature information need only be stored at the highest possible level of abstraction, maximum elegance and economy of storage is achieved. These formalisms, however, are not easily amenable to representing uncertainty in an elegant and efficient manner.

In reasoning and decision making under uncertainty, specialized graph-based formalisms like influence diagrams (Howard & Matheson, 1981) have been advocated for computer representation of probabilistic knowledge and decision models. These formalisms focus on the dependencies among the probabilistic variables. They encode probabilistic models as directed graphs with the nodes representing the uncertain variables and the directed arcs denote possible probabilistic dependence between variables. Each node encodes a conditional probability distribution of that node's variable given each combination of values of its direct predecessors nodes. Various techniques have been developed over the last decade for probabilistic inference and reasoning using this representation; see for example Pearl (1988).

Pc-nets combine the formalisms of influence diagrams with inheritance hierarchies by representing concepts with influence diagrams and then linking these conceptual diagrams in a hierarchy. By do so, pc-nets are able to capture the best features of both formalisms and to use them effectively in support of utility-based categorization.

## 3 An Application in Automated Machining

We will illustrate the use of pc-nets in utility-based categorization with a real-world example of an automated machining problem. This is similar to an application described by Agogino and Ramamurthi (1990). Unattended or automated machining operations are important parts of any intelligent manufacturing system. It requires the automation of the human operator's efforts to monitor and make appropriate adjustments to the state of the machine. An automated machining system typically has sensors which acquire data on (1) dimensions of the workpiece, (2) acoustic emission from the machining processes, (3) cutting forces (dynamometer readings), and (4) electric current (ammeter), etc. These data are then used to determine the state of the machine and appropriate action or actions are taken to ensure the continuous operation of the plant so as to minimize production cost, thereby maintaining competitiveness. The possible states of the machining process at various level of abstraction are illustrated in Figure 2.

At the most abstract level, the state of the machining process is either "within variability limits" or "out of variability limits." Refining the concept "out of variability limits" are "tool failure," "sensor failure" and "transient state." The latter occurs during entry or exit of the cutting tool into the workpiece. Refining "tool failure" are "tool chatter" which is typically characterized by an event in which an acoustic emission signal increases dramatically in amplitude as does the frequency content of the dynamometer. If left unchecked, tool chatter can result in tool, workpiece or machine damage. Remedies for this problem include archies arose in discussions with Eric Horvitz.

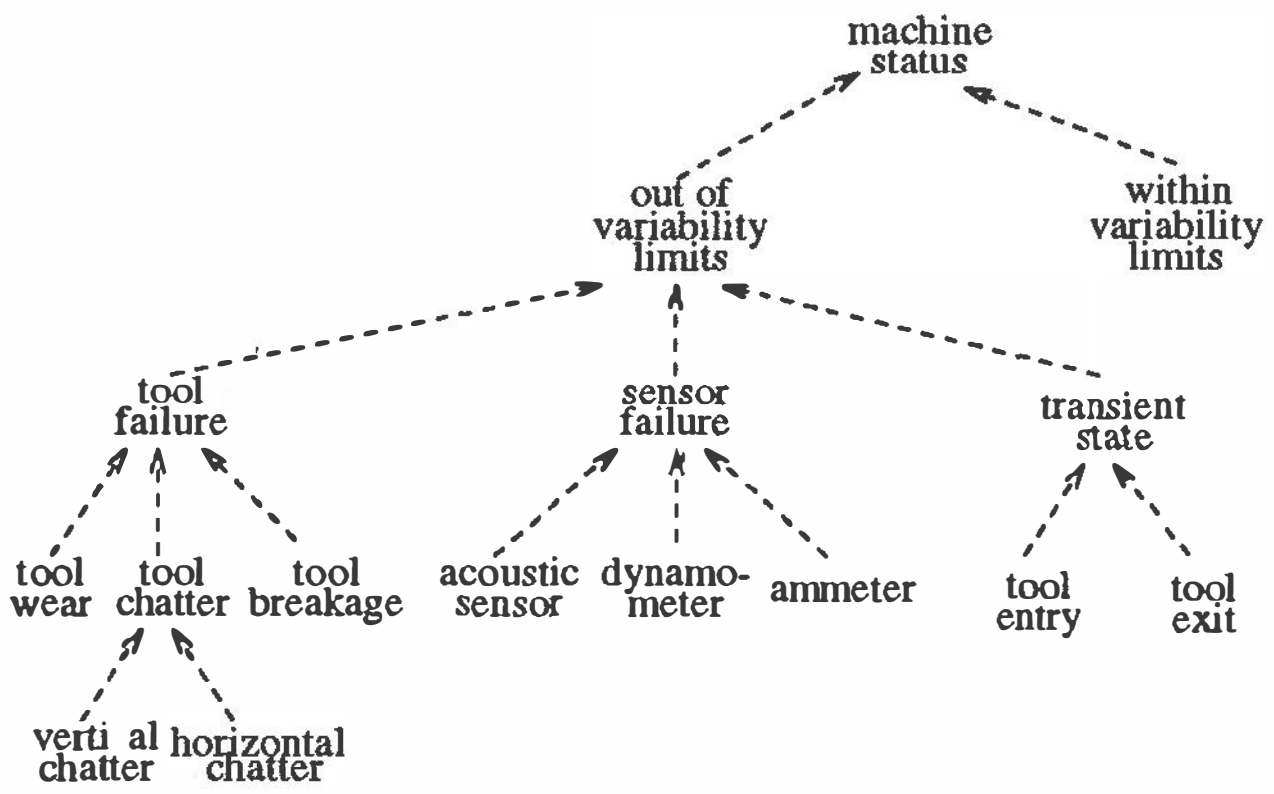

Figure 2: Hierarchy for states of a machine

reducing the depth of cut or reducing the feed rate. "Tool wear" is typically characterized by a gradual increase in acoustic emissions, and by a gradual increase in cutting force as measured by the dynamometer. A tool that is worn out needs to be re-sharpened or replaced in order to achieve the desired surface finish and dimensional tolerances. "Tool breakage" is typically characterized by an acoustic emission exhibiting a high amplitude peak at the moment of tool fracture, and followed by a sharp drop in signal amplitude to a level below that of normal. It is also characterized by a large rise in cutting forces, followed by a drop before finally continuing at a value above the average. Tools that are broken cannot perform any machining task and must be replaced immediately.

This problem is interesting because under differing operating conditions and situations, different levels of abstraction in monitoring and reasoning may be desired. For example, if the tool has been changed recently, giving it a low prior probability that it will fail soon, it may be more worthwhile to only monitor at a more general level, i.e., "tool failure," "sensor failure" and "transient state," rather than spent extra resources to differentiate the finer details. In other words, the expected value of the information derived from using more detailed concepts may not justify the required additional computational resources. On the other hand, if the tool has already been in used for a long time, then it might be worth the extra effort spent in monitoring and reasoning with more detailed concepts, like for example at the level of "tool chatter," "tool wear," "tool breakage," "sensor failure" and "transient state." Also if the material currently being machined is a difficult one, e.g., a high-carbon steel, which is known to have caused occasional tool breakage, then it may also be worthwhile to monitor at a deeper level of detail. In another possible scenario, suppose the some critical sensors are out of order, then the only level of detail available might be at the most abstract level whereby only two possible states are being monitored. The operator would then need to be alerted to take any corrective action.

# 4 Probabilistic Conceptual Network

## 4.1 Definitions

A *probabilistic conceptual network* (*pc-net*) consists of a *probabilistic concept hierarchy* (*pc-hierarchy*) connecting a set of *probabilistic concept diagrams* (*pc-diagrams*). Each node in the pc-hierarchy represents a concept, and the links in the hierarchy specify subsumption relations among the concepts thereby organizing the concepts at various level of abstraction or specificity. Associated with each subsumption link is a value indicating the conditional probability that a concept holds given that its immediate superconcept holds. Individually, each concept within the pc-hierarchy is represented by a pc-diagram. We may visualize a pc-net as a hierarchical organization of pc diagrams.

A pc-diagram for a concept is a special probabilistic influence diagram (pid)[2] representing the knowledge about the probabilistic relations between the concept and the features that characterize it. The concept is represented as a deterministic node while the features are represented by chance nodes. As a convention, we direct arcs by default, from the concept to its feature nodes. For each feature node $F$ in the pc-diagram for concept $C_k$, we store a probability distribution of the form

$$p(F|C_k, B^k(F))$$

where $B^k(F)$ is the set of conditional predecessors (possibly empty) that excludes $C_k$. We shall assume that background information $\xi$ is used in all the probability distributions. We represent $C_k$ as a deterministic node because we do not need the distribution $p(F|\neg C_k, B^k(F))$. A pc-diagram for a concept provides information for discriminating that concept from other concepts in a domain. Pc-diagrams allow knowledge to be represented locally providing modularity in the knowledge-base.

The value of a pc-net emanates from its ability to support utility-based categorization. As shown previously in Figure 1, given a pc-net together, a conceptual cover can be selected at some level of abstraction to construct a categorization decision model corresponding to that level of abstraction. We shall describe the procedures for model construction in Section 5.

Finally, pc-net uses an inheritance mechanism whereby a concept may share information about features from a concept higher up the hierarchy. It does so by taking advantage of a form of conditional independence called *subconcept independence*[3] which is not conveniently represented in ordinary influence diagram representation. A feature is said to be subconcept independent of a concept if knowledge about the feature does not affect the agent's belief about any of that concept's subconcept. We will have more to say about subconcept-independence in Section 4.5.

[2] A pid is an influence diagram with only probabilistic nodes and conditioning arcs.

[3] Section 6.1 compares subconcept independence with "subset-independence" in similarity networks.

| Feature | Description |
|---|---|
| AE-mag | acoustic emission magnitude |
| ΔAE-mag | change in acoustic emission magnitude |
| AE-freq | acoustic emission frequency |
| dyn-freq-x | cutting force frequency in x-direction |
| dyn-freq-y | cutting force frequency in y-direction |
| AE-mean | mean of the acoustic signal |
| ΔAE-mean | change in the mean of the acoustic signal |
| dyn-rms-x | cutting force in the x-direction |
| Δdyn-rms-x | change in cutting force in the x-direction |
| dyn-rms-y | cutting force in the y-direction |
| Δdyn-rms-y | change in cutting force in the y-direction |
| AE-peak | acoustic emission peak value |
| dyn-peak-x | peak cutting force in x-direction |
| dyn-peak-y | peak cutting force in y-direction |
| current | motor current |

Table 1: Descriptions of features

## 4.2 Automated Manufacturing Example

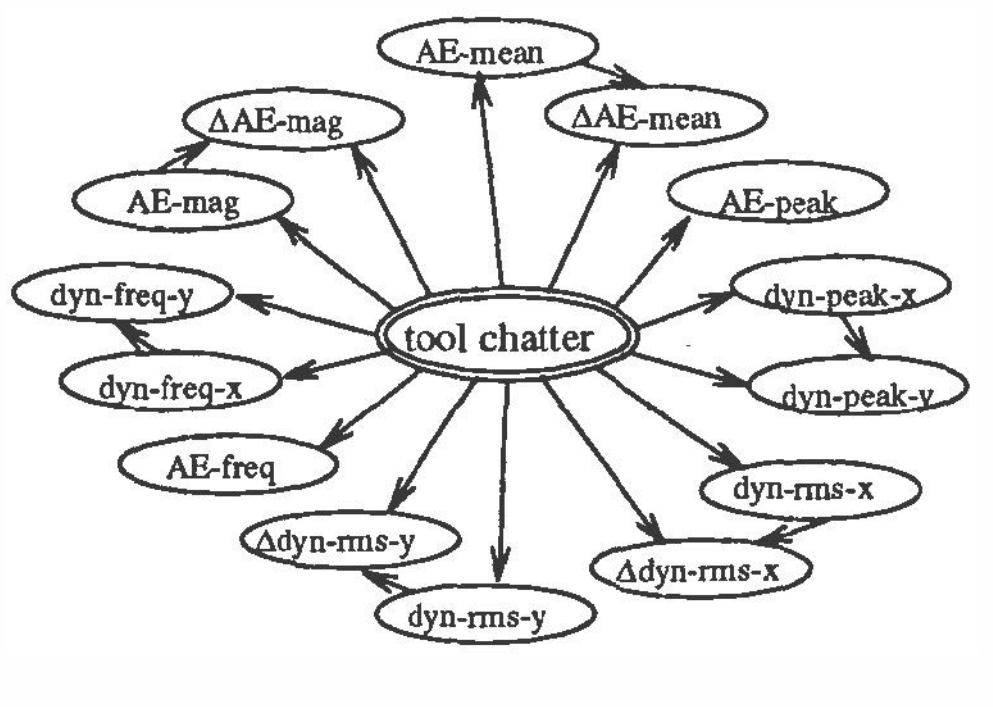

Figure 3: The pc-diagram for "tool chatter."

Figure 3 shows the pc-diagram "tool chatter." This diagram comprises a deterministic node representing "tool chatter" and a number of feature nodes whose descriptions are given Table 1. An arc between two feature nodes indicates that these two features may not be conditionally independent given the concept "tool chatter." For example, the arc between the node "AE-mag" and the node "ΔAE-mag" indicates that information about the current magnitude of acoustic emission may provides information about the change in magnitude of acoustic emission. The direction of this arc could be reversed without any change in assertion about possible dependency.

Figure 4 shows a fragment of the full pc-net for the automated machining showing the concepts "tool failure," "tool chatter," "tool wear," "tool breakage," "sensor failure," and "transient state."

## 4.3 Probabilistic Subsumption Relations

We shall denote the fact $C_i$ is a subconcept of $C_j$ by $C_i \ll C_j$. The set of the most general subsumees (i.e. all the direct subconcepts) of $C_k$ is denote by $\ll(C_k)$, and the most specific subsumer of a set of concepts $S = \{C_1, \ldots, C_n\}$ is denoted by $\gg(C_1, \ldots, C_n)$ or $\gg(S)$.

Suppose $C_i \ll C_j$. We have $p(C_i|C_j) = \frac{p(C_i \wedge C_j, \xi)}{p(C_j)}$. But $C_i \ll C_j$ implies that $p(C_i \wedge C_j) = p(C_j)$. Therefore

$$p(C_j|C_i) = \frac{p(C_j)}{p(C_i)}. \quad (1)$$

In other words the subsumption probability is simply the ratio of the prior probabilities of the concepts it connects.

## 4.4 Feature Relations and Conceptual Abstraction

Suppose we have already assessed a set of pc-diagrams, we can combine them to produce a more general super-concept. For example, given the pc-diagrams for "tool chatter," "tool wear," and "tool breakage," we can obtain the pc-diagram for "tool failure." In general, given $C_k$, set of its most general subsumees $\ll(C_k)$, and suppose $B^k(F) = \cup_{C_j \in \ll(C_k)} B^j(F)$ then $p(F|C_k, B^k(F)) = \sum_{C_j \in \ll(C_k)} p(F|C_j, C_k, B^k(F)) p(C_j|C_k, B^k(F))$. The feature $F$ is independent of $C_k$ given any subconcept $C_j$ of $C_k$ since once $C_j$ is known to be true then any information about $C_k$ will not have any further effect on our belief on $F$. This implies that $(F^i|C_j, C_k, B^k(F)) = p(F^i|C_j, B^k(F))$. Likewise, $C_j$ is independent of $B^k(F)$ given $C_k$. Hence $p(F|C_k, B^k(F)) =$

$$\sum_{C_j \in \ll(C_k)} p(F|C_j, B^k(F)) p(C_j|C_k),$$

which may be rewritten as

$$\sum_{C_j \in \ll(C_k)} p(F|C_j, B^j(F), B^k(F) \setminus B^j(F)) p(C_j|C_k).$$

But the set of conditioning features $B^k(F) \setminus B^j(F)$ is independent of $F$ given $C_j$. Hence $p(F|C_k, B^k(F)) =$

$$\sum_{C_j \in \ll(C_k)} p(C_j|C_k) p(F|C_j, B^j(F)).$$

Hence if $C_k$ is a concept in the pc-net and all the pc-diagrams for the concepts in $\ll(C_k)$ has been assessed, then the pc-diagram for $C_k$ may be derived from those of its subconcepts. Formally, for any feature $F$, $p(F|C_k, B^k(F))$

$$= \sum_{C_j \in \ll(C_k)} p(C_j|C_k) p(F|C_j, B^j(F)) \quad (2)$$

where $B^k(F) = \cup_{C_j \in \ll(C_k)} B^j(F)$

Equation (2) allows us to build the pc-net from bottom up by propagating the probability distributions in the pc-diagrams from the bottom of the hierarchy up to the root of the hierarchy. This allows us to build the pc-net by first constructing the pc-diagrams for all the terminal or atomic concepts, and then the pc-diagrams for the more general concepts may be derived from the pc-diagrams below them. However, it is possible to simplify the pc-net by identifying subconcept independence and take advantage of inheritance.

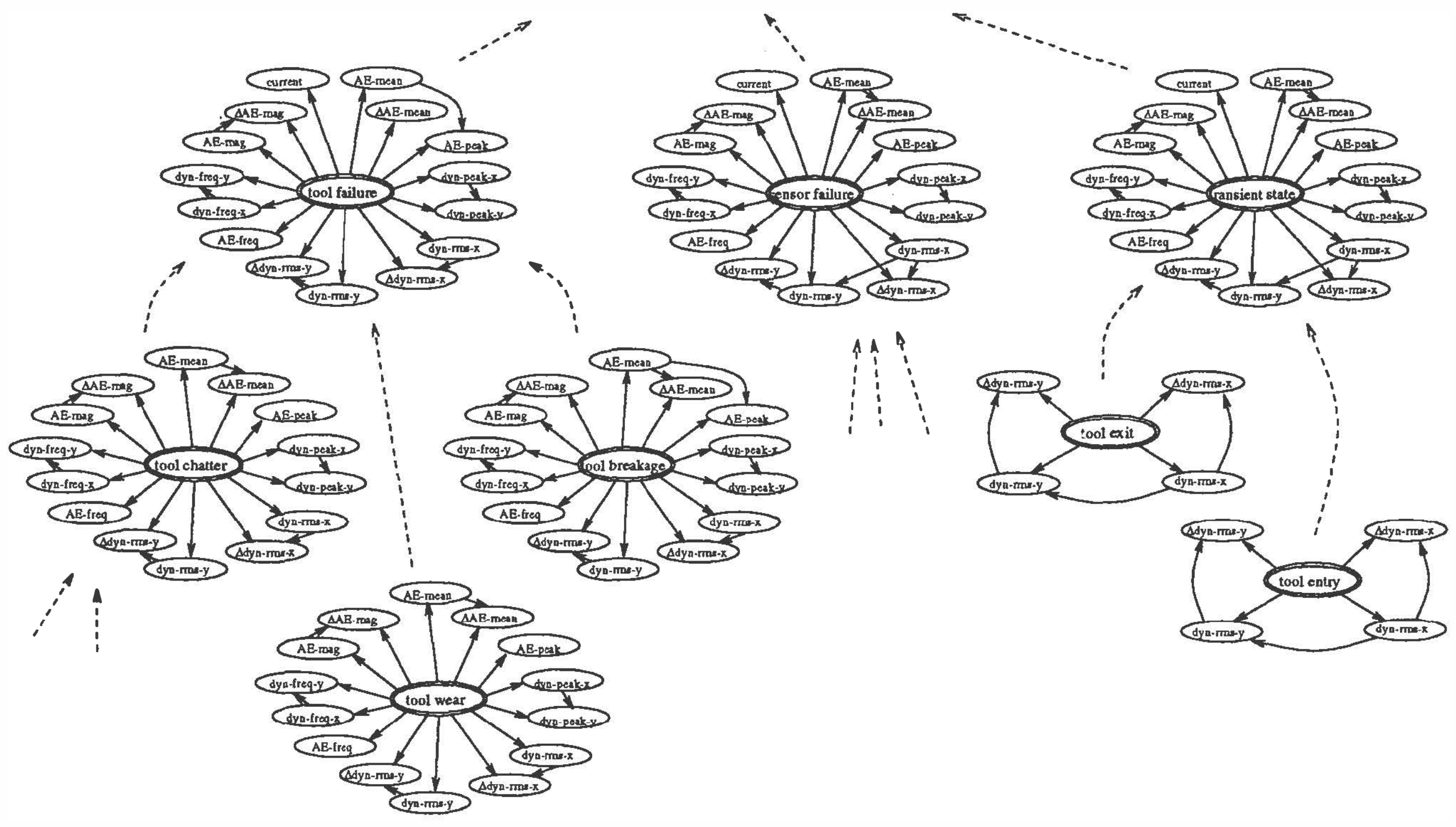

Figure 4: A fragment of the pc-net for the automated machining problem

### 4.5 Feature Inheritance for Subconcept-Independent Concepts

The principle of inheritance in pc-net is based on a special type of independence that can hold among concepts and features. Formally, we say that a feature $F$ is ***subconcept independent*** of a concept $C_k$ given $B$, if and only if

$$p(C_i|f, C_k, B) = p(C_i|C_k, B) \qquad (3)$$

for all feature values $f$ of $F$ and for all subconcepts $C_i$ of $C_k$. Intuitively, information about a feature that is subconcept independent of a concept does not affect the agent's belief about any of that concept's subconcepts. An equivalent criterion for subconcept independence is obtained using using Bayes' rule:

$$p(F|C_i) = P(F|C_k) \quad \forall C_i \ll C_k. \qquad (4)$$

The last equation applies that for any pair of subconcepts $C_i$ and $C_j$ of $C_k$, i.e., $p(F|C_i) = p(F|C_j)$. Conversely, if the last equation holds then using equation (2), $p(F|C_k) = \sum_j p(C_j|C_k)p(F|C_j) = \sum_j p(C_j|C_k)p(F|C_i) = p(F|C_i)\sum_j p(C_j|C_k) = p(F|C_i)$. Hence an equivalent criterion for subconcept independence is:

$$p(F|C_i) = P(F|C_j) \quad \forall C_i, C_j \ll C_k. \qquad (5)$$

We shall denote by $F \perp_{\ll} C_k|B$, the fact that $F$ is subconcept independent of $C_k$ given $B$. In cases where the background knowledge is understood, the $B$ may be omitted. An interesting property about $\perp_{\ll}$ is that once it has been established for a concept, it recursively applies to all of its subconcepts (Poh, 1993). That is,

$$F \perp_{\ll} C_i|B \Longrightarrow F \perp_{\ll} C_j|B \quad \forall C_j \ll C_i. \qquad (6)$$

The justification for the application of inheritance for subconcept independent concepts for a feature is due to equations (4) and (5). Since the probability distributions for the feature are identical, we need only store them at the highest possible position.

To illustrate the idea of inheritance, consider the fragment of the pc-net for "transient state," "tool exit" and "tool entry" shown in Figure 4. The feature "$\Delta$rms current" is subconcept independent of "transient state." We do not need to explicitly store the probability distributions for "$\Delta$rms current" in the pc-diagrams for "tool entry" and "tool exit." That is, we may "omit" these probability distributions (and hence the corresponding feature nodes) in their respective pc-diagrams. When needed, the probability values are filled in by inheriting them from "transient state."

# 5 Model Construction

## 5.1 Constructing Categorization Decision Models

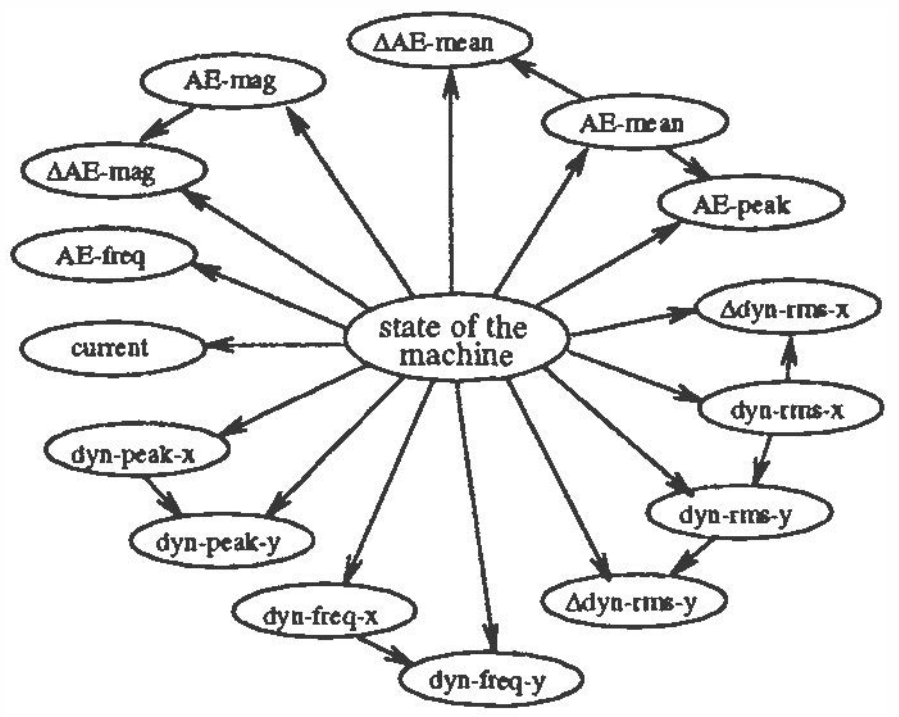

Figure 5: The categorization prob. influence diagram

We shall illustrate how a categorization decision model may be constructed from the pc-net for the automated machining problem. In this application, the preference model may be expressed in the form $v(A_k, C_i)$ where $A_k$ is an action that may be taken, like for example, "reducing cutting speed", "reducing depth of cut", etc. $C_i$ is any state of the machining operation we have described earlier. $v(A_k, C_i)$ gives the utility of the outcome by taking action $A_k$ when the state of the machining operation is $C_i$.

Suppose the sensors report information on "AE-mag," "AE-rms," "dyn-rms-x," "dyn-rms-y," and "rms-current," and our utility-based categorical reasoner described earlier, determines that the most appropriate level of abstraction corresponds to the set of concepts comprising "tool chatter," "tool wear," "tool breakage," "sensor failure" and "transient state." We can combine the respective pc-diagrams for these five concepts to construct a *categorization probabilistic influence diagram* as shown in Figure 5. The graphical structure of the combined categorization influence diagram is obtained by performing graphical union of the individual pc-diagrams while treating each central concept node as being the same node in each of the individual pc-diagrams. Notice that the concept node in the constructed diagram is now a probabilistic variable ($\mathcal{C}$) ranging over the five concepts used in its construction. The conditional probabilities for each of the feature nodes in the constructed diagram is obtain by copying over their respective original values in the individual pc-diagrams. That is, for any feature $F$,

$$p(F|\mathcal{C} = C_i, B^g(F)) = p(F_i|C_i, B^i(F)) \qquad (7)$$

where $B^g(F)$ is the set conditional predecessors of $F$ excluding $\mathcal{C}$, in the constructed diagram.

The next step in the construction procedure is to complete the diagram by turning it into a categorization decision model as shown in Figure 6. This is done by first, adding the decision and value node to reflect the preference model described earlier. Next, informational arcs from the observed feature nodes to the decision node are added. The completed categorization influence diagram can now be solved using existing methods (Shachter, 1986).

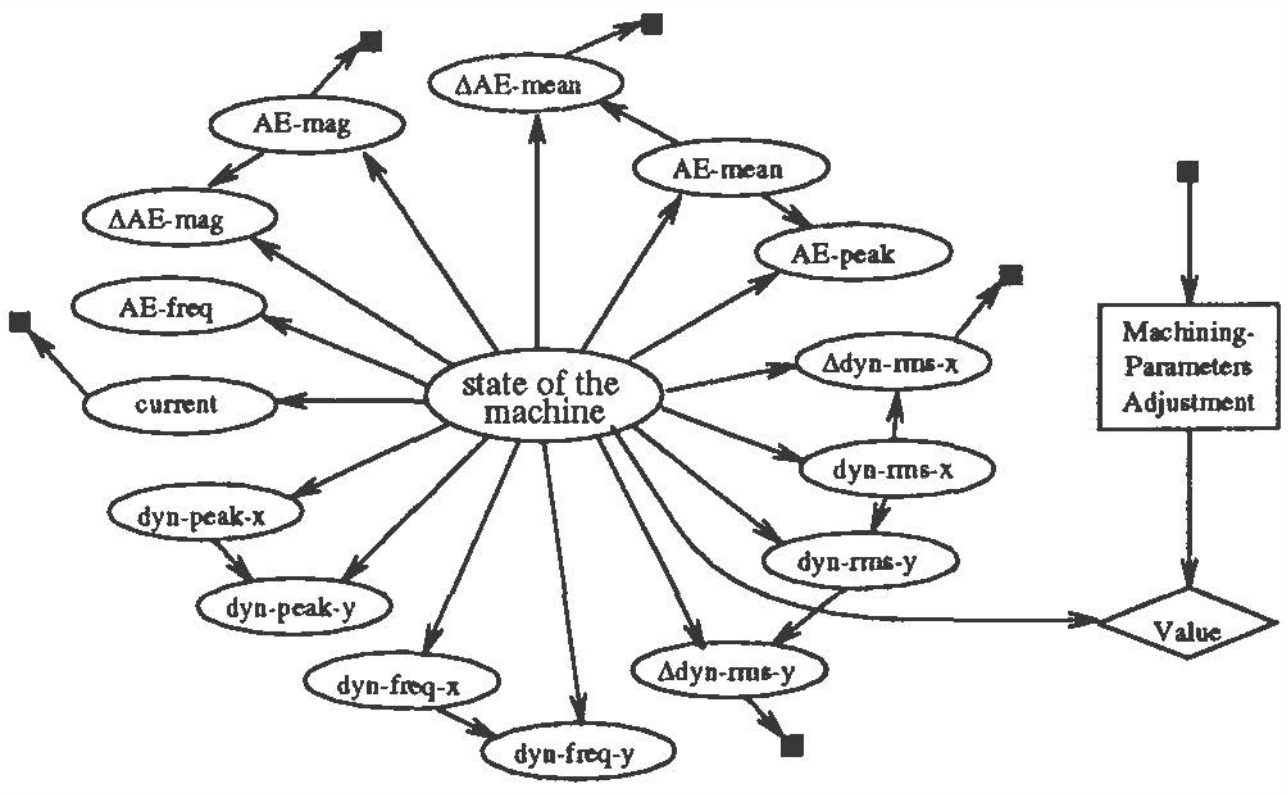

Figure 6: The categorization decision model

## 5.2 Validity of the Constructions

An important characteristic of our decision model construction procedure is that the final model so constructed must reflect as accurately as possible the state of information originally asserted by the knowledge-base and preference model. Our knowledge-base contains assertions about concepts, their properties, and the probabilistic relationships among them. Validity of a probabilistic model construction depends on the soundness of the construction procedure. Heckerman (1991) suggests that soundness should be characterized by the preservation of the joint-distribution of the variables involved across the construction. For pc-net, it can be shown that if the pc-diagrams in a given conceptual cover are mutually *consistent*, then the construction is indeed sound (Poh, 1993).

# 6 Related Work

## 6.1 Probabilistic Similarity Networks

Probabilistic similarity network (Heckerman, 1991) is a knowledge engineering tool for building probabilistic influence diagrams. We shall briefly describe the similarities and differences between pc-net and similarity network here. A more comprehensive comparison is available in (Poh, 1993). Both pc-net and similarity networks are capable of building the same type of influence diagrams, but pc-net is able to do so at varying levels of abstraction, whereas similarity network can only do so at one level. Another major difference is that pc-net is capable of representing categorical abstraction relations whereas similarity networks can't. Another difference is that the probabilities in a pc-net are assessed before categorical reasoning and model construction take place whereas in similarity networks, all the knowledge maps are initially unassessed and are carried out only after the global knowledge map has been built.

Both pc-net and similarity network use some sort of local influence diagrams for concept representation. However, a local knowledge map in similarity network is built based on a pair of concepts. There are also differences between a pc-diagram and a hypothesis-specific knowledge map (hs-map) in similarity networks. First, the concept node is included in the pc-diagram, whereas, it is not part of the hs-map. Second, a pc-diagram is always a connected graph whereas a hs-map may not be. Finally, a pc-diagram has its probabilities initially assessed whereas, a hs-map is not.

The notion of subconcept independence in a pc-net is analogous to *subset independence* used in conjunction with *partitions* in similarity networks. Similarity networks use partitions to speed up assessment while pc-net saves assessments and storage by using inheritance mechanisms based on subconcept independence. In pc-net terms, a partition for a feature in similarity network can be viewed as an an abstracted concept subsuming all the concepts in the partition. Furthermore, that feature is subconcept independent of the abstracted concepts. Assessing the probability distribution for the feature given the abstracted concept and applying inheritance is equivalent to assessing the probabilities within the partition.

### 6.2 Knowledge-Based Model Construction Methods

Several approaches have been proposed for construction or building of influence diagrams. There approaches may be classified under two highly contrasting methodologies. The first, known as the *synthetic* approach (Horvitz, 1991) starts with the empty influence diagram; nodes and arcs are added to the model through some methods of inference based on simple rules or relationships. These inferences are usually driven by assertions about the world, goals, or utility (Holtzman, 1989; Breese, 1987; Goldman & Charniak, 1990; Wellman, 1988). These approaches however, usually do not have principled control over the degree of abstraction or details in the model that they are building other than using some heuristics. The second, known as the *reduction* approach (Horvitz, 1991) seeks to custom-tailor comprehensive, intractable decision problems to specific challenges at run time through a pruning procedure that removes irrelevant distinctions and dependencies (Heckerman & Horvitz, 1991).

The decision model construction approach based on probabilistic conceptual networks developed in our research does not commit to either of these two contrasting approaches, but instead, employs mixed strategies. The approach can be seen as synthetic to some extent in that it builds an influence diagram dynamically at runtime. However, unlike the pure synthetic approaches, the building blocks used by this approach are not individual nodes and arcs, but rather modules of localized influence diagrams. On the other hand, the approach can be seen to be reducible in that modules of local influence diagram have been pre-assessed. However, instead of pre-assessing a comprehensive influence diagram, pc-net does not commit to one large influence diagram, but instead, is a comprehensive networks of related local probabilistic influence diagrams. The approach here allows for reasoning about the relationship among these local influence diagrams, and combines only those that are relevant or are required while discarding those not required in the decision model it is building.

The advantage of our approach over that of the comprehensive model reduction approach, is that assessing smaller and more focused local pc-diagrams is usually easier and more manageable as compared with attempting to assess a huge comprehensive influence diagram. This local-to-global approach to constructing large probabilistic influence diagrams has been demonstrated with similarity networks.

The advantage of this approach over that of the complete synthetic approach is that the construction procedure is controlled using well founded principles of decision theory. We use a principled approach to reason about the values of constructing different parts of the model. The model being built can be custom-tailored to the optimal level of abstraction and avoid any unnecessary details. This is very important when we consider computational or resource constraints.

## 7 Conclusion

Previous work on integrating uncertainty and categorical knowledge representation has been done with a broad range of emphases and purposes. Saffiotti (1990) proposed a general framework for integrating categorical and uncertainty knowledge. In particular, Shastri (1985) proposed a semantic-network-like representation language for evidential reasoning using the principle of maximum entropy. Similarly, Lin and Goebel (1990) proposed a graphical scheme integrating probabilistic, causal and taxonomic knowledge for abductive diagnostic reasoning. This latter formalism has two types of links, namely "is-a" and "causal." In classifier-based reasoning, term subsumption languages are being extended to accommodate plausible inferences (Yen & Bonisson, 1990). More recently, Leong (1992) proposed a network formalism using various kinds of links including "a kind of," temporal precedence, qualitative probabilistic influence (Wellman, 1988) and property relations ("Context"). Many of these formalisms have desirable features that we need, but none has all.

Finally, by combining the formalisms of influence diagrams and abstraction hierarchies, pc-nets effectively represent both categorical knowledge/relations and uncertainty in a modular and compact way. It can also support dynamic construction of a specific class of decision model at varying levels of abstraction. We have also demonstrated the applicability of pc-net to real-world applications in automated machining.

## Acknowledgements

We wish to thank Ross Shachter, Eric Horvitz and the anonymous referees for their helpful comments and suggestions on the content of this paper.